Running Head:  Simplicity-based approach to language learning

Language learning from positive evidence, reconsidered: A simplicity-based approach


*Anne S. Hsu*

Department of Cognitive, Perceptual and Brain Sciences

University College London

26 Bedford Way

London, WC1H 0AP, UK

anne.hsu@ ucl.ac.uk

*Nick Chater*

Behavioural Science Group

Warwick Business School

University of Warwick

Coventry, CV4 7AL, UK

Nick.Chater@wbs.ac.uk

*Paul Vitányi*

Centrum Wiskunde & Informatica,

Science Park 123, 1098 XG Amsterdam,

The Netherlands

paul.vitanyi@cwi.nl







*Abstract*

Children learn their native language by exposure to their linguistic and communicative environment, but apparently without requiring that their mistakes are corrected. Such learning from "positive evidence" has been viewed as raising "logical" problems for language acquisition. In particular, without correction, how is the child to recover from conjecturing an over-general grammar, which will be consistent with any sentence that the child hears? There have been many proposals concerning how this "logical problem" can be dissolved. Here, we review recent formal results showing that the learner has sufficient data to learn successfully from positive evidence, if it favours the *simplest* encoding of the linguistic input. Results include the learnability of linguistic prediction, grammaticality judgments, language production, and form-meaning mappings. The simplicity approach can also be "scaled-down" to analyse the learnability of specific linguistic constructions, and is amenable to empirical test as a framework for describing human language acquisition.




Children appear to learn language primarily by exposure to the language of others. But how is this possible? The computational challenges of inferring the structure of language from mere exposure are formidable. In light of this, many theorists have conjectured that language acquisition is only possible because the child possesses cognitive machinery that fits especially closely with the structure of natural language. This could be because the brain has adapted to language (Pinker & Bloom, 1990), or because language has been shaped by the brain Christiansen & Chater, 2007).

A number of informal arguments concerning the challenge of language learning from experience have been influential. Chomsky (1980) argued that the "poverty of the stimulus" available to the child was sufficiently great that the acquisition of language should be viewed as analogous to the growth of an organ, such as the lung or the heart, unfolding along channels pre-specified in the genome. Here, we focus on a specific facet of poverty of the stimulus: that children do not appear to receive or attend to "negative evidence:" explicit feedback that certain utterances are ungrammatical (Bowerman, 1988; Brown & Hanlon, 1970; Marcus, 1993).[1]

The ability to learn language in the absence of negative evidence is especially puzzling, given that linguistic rules are riddled with apparently capricious restrictions. For example, a child might naturally conclude from experience that there is a general rule that *is* can be contracted, as in *He's taller than she is*. But contractions are not always allowed, for example: *\*He is taller than she's*. The puzzle is that, once the learner has entertained the possibility that the overgeneral rule is correct, it appears to have no way to "recover" from overgeneralization and recognise that restrictions should be added. This is because each contraction that it hears



conforms to the overgeneral rule. Now, if the learner uses the overgeneral rule to *generate* language, then it will from time to time produce utterances such as *John isn't coming but Mary's*. A listener's startled reaction or look of incomprehension might provide a crucial clue that the rule is overgeneral. However, such feedback is the very negative evidence that appears to be inessential to child language acquisition. Thus, if children do not require such negative evidence, how can they recover from such overgeneralisations? Various scholars argue that they cannot: Restrictions on overgeneral grammatical rules must, instead, be innately specified (e.g., Crain & Lillo-Martin, 1999). Other theorists argue that avoiding overgeneral rules poses a fundamental "logical problem" for language acquisition (Baker & McCarthy, 1981; Dresher & Hornstein, 1976).

One way to defuse the puzzle is to challenge its premise. One possibility is that, despite appearances, children can access and use negative evidence in a subtle form. In this paper, we set aside these contentious issues (e.g., Demetras, Post & Snow, 1986; Marcus, 1993) and argue that, whether or not negative evidence is available to, or used by, the child, language can successfully be learned without it (following, for example, MacWhinney,  1993; 2004; Rohde & Plaut,  1999; Tomasello,  2004).

The arguments for learnability from positive evidence presented here are part of a broader tradition of research on learnability (e.g., Angluin, 1980, 1988; Clark, & Eyraud, 2007; Feldman, 1972; Gold, 1967; Horning, 1969; Jain, Osherson, Royer & Kumar Sharma, 1999; Niyogi, 2006; Wharton, 1974). And formal learnability arguments are complementary to recent developments of language engineering systems, which have shown that it is possible to learn automatically non-trivial aspects of phonology, morphology, syntax and semantics from positive language input (Goldsmith, 2001; Klein & Manning, 2005; Steyvers, Griffiths &



Tenenbaum, 2006). While such systems are still very far from being able to acquire language from mere exposure, the pace of progress suggests that *a priori* barriers to learning may not necessarily be insurmountable.

Rather than surveying these developments, and indicating how they may be extended, here we will take a more direct approach: we focus on one major line of positive learnability results based on the 'simplicity principle'. We begin by introducing the simplicity principle (Section 1) and considering how it can be embodied in an "ideal learner" (Section 2). We then outline some recent formal results on how the simplicity principle can be used to learn aspects of language such as utterance prediction, grammaticality judgments, language production, and mapping between form and meaning (Sections 3-6). We then briefly describe a practical method for assessing learnability of linguistic patterns using the simplicity approach, and how this assessment can be linked with experimental data (Section 7). Overall, the contribution of the work reviewed here is to show that, under fairly mild conditions, language acquisition from sufficient amounts of positive evidence is possible; and to indicate how the simplicity-based approach can potentially provide a framework for understanding child language acquisition.

## *1. Ideal learning using a simplicity principle*

The simplicity principle has a long history in the philosophy of science and the study of perception (e.g., Mach, 1959/1886), and has been proposed as a general cognitive principle (Chater & Vitányi, 2002). A formal analysis of simplicity learning starts with supposing a learner (human or artificial) that is faced with a set of positive data. For language, this data is a set of observed grammatical sentences.[2] Any set of observed sentences will be consistent with an infinite number of grammars. That is, any set of sentences could have been generated by any



of an infinite number of grammars. How can the learner choose among these infinite possibilities?

The simplicity principle recommends that the learner prefers hypotheses which allows for the *simplest* encoding of the data. For language, the data will be the observed sentences, and hypotheses are grammars (or other linguistic representations), which can be viewed as a set of probabilistic rules which captures the patterns in the linguistic input to the learner. Simplicity can be measured by viewing hypotheses (here, grammars) as *computer programs* which *encode* the data (the data is generated as the *output* of the program). The simplicity principle thus favors the grammar that provides the shortest encoding of the data.[3]

How can a grammar be viewed as a computer program for encoding linguistic input? One concrete approach involve two steps. The first step is to specifying the grammatical rules (and, crucially, probabilities of their use). This defines a probabilistic process for generating sentences; and thus defines a probability distribution over possible strings. The second step is to encode the specific sentences in the input. It is intuitively clear that the most efficient way to do this is to reserve shorter codes for probable strings; and longer codes for less probable strings. A basic result from information theory (e.g., Cover & Thomas, 2006) is that the optimal way to do this is to assign a binary code of length $\log_2 1/p$ to a string with probability $p$.[4] So, intuitively, a (probabilistic) grammar provides a short encoding of the linguistic input if it can itself be specified briefly; and if it makes the sentences that are actually observed as probable as possible. There is a tension between these objectives. An "over-precise" grammar, which encodes exactly those sentences that have been encountered and no others will make those particular sentences highly probably; but the code for such a grammar will be long (roughly, it will consist just of a list of the "allowed" sentences). Conversely, a very simple but



overgeneral grammar (e.g., stating, roughly, that words can occur in any order with equal probability) will have a short code, but, because the space of possible allowed sentences is vast, the code for the specific sentences observed by the learner will be very long. The simplicity principle recommends finding an balance between these extremes: postulating restrictions in the grammar just when these "pay off" by sufficiently reducing the code length of the sentences, encoded by the grammar.

As we have indicated, in general, the better the grammar captures the structure of the language, the shorter the encoded representation of the linguistic input will be. For a concrete example, let us first consider hypotheses (i.e. grammars) describing artificially simplistic language data. Suppose the observed language was the following repeating infinite string of sentences: Hi! Bye! Hi! Bye!... One hypothesis could be "The language is a sequence of 'Hi!' and 'Bye!' occurring independently, and each with .5 probability." Under this hypothesized grammar, the encoded specification of the language input will be "0101…", where 0 and 1 correspond to 'Hi!' and 'Bye!' respectively. Now if the hypothesis was a more powerfully descriptive grammar such as "The language contains a single sentence 'Hi! Bye!', then no further code at all is required to specify the linguistic input. Now, an infinite language input is fully specified in a simple finite description—and, more generally, the more precisely the grammar captures the structure of the linguistic input, the shorter the encoding of that linguistic input will be.

Initially, the learner may not have sufficient data to favour the latter hypothesis; but eventually the latter "grammar" will provide the simpler encoding, because it correctly captures regularities in the input. Hence, as linguistic input accumulates, the grammar which provides the simplest encoding will be updated. An *ideal* simplicity learner (as in the mathematical



results below) will have access to all (infinite) possibly hypothetical grammars that describe its current language data input, and choose the "simplest;' any real, and hence computationally limited, learner can of course only approximate this calculation to some degree.

Crucially, note that the simplicity-based learner has a mechanism for avoiding overgeneral grammars, when learning from positive evidence. Although our artificial data is compatible with a random sequence of 'Hi!' and 'Bye!,' the corresponding grammar is eliminated without the need for negative evidence, but because another grammar provides a shorter encoding of the input.[5]

This point applies equally to learning natural languages. Consider the case of *is* contraction mentioned above. Consider two possible grammars, one that allows *is* contraction everywhere, and one that is more restricted (allowing *He's taller than she is* but not *\*He is taller than she's*). The latter "grammar" will be more complex (because it involves specifying more precisely when contraction can occur); but it will encode the linguistic input more briefly, because it more accurately captures the structure of the language. Given sufficient linguistic input, the benefit of the more accurate encoding of the linguistic input will overwhelm any additional costs in encoding the grammatical rule, and the more precise rule will be favoured. Thus, it appears that an overgeneral grammar can be eliminated by applying the simplicity principle to positive data only.

This intuition is encouraging but hardly definitive. Knowing that a learner can potentially eliminate a single over-general grammar does not, of course, indicate that it can successfully choose between an infinity of possible grammars, and home in on the "true" grammar, or some approximation to it. We shall see, however, that positive mathematical results along these lines are possible. Moreover, in Section 7, we shall apply the style of



argument sketched above to the learnability of some specific, and much-discussed, linguistic regularities (see Hsu, Chater & Vitányi, 2011).

<p style="text-align:center;">*2 An "ideal" learner*</p>

Below, we consider some formal theoretical results describing what an "ideal" learner can learn purely from exposure to an (indefinitely long) sequence of linguistic input (i.e., positive evidence) by using the simplicity principle.

What is the structure of the linguistic material to be learned? Fortunately, it turns out that we need assume only that this input is generated probabilistically by some computable process.[6] This restriction is mild because cognitive science takes computability constraints on mental processes, including the generation of language, as founding assumptions (Pylyshyn, 1984) and, indeed, specific models of language structure and generation all adhere to this assumption. Finally, for mathematical convenience, and without loss of generality, we assume that the linguistic input is coded in binary form.

Importantly, note that these assumptions allow that there can be any (computable) relationship between different parts of the input---we do not, for example, assume that sentences are independently sampled from a specific probability distribution. Our very mild assumption allows sentences to be highly interdependent (this is one generalization with respect to earlier results, e.g., Angluin, 1980; Jerome Feldman, 1972; Wharton, 1974), and includes the possibility that the language may be modified or switched during the input or indeed that sentences from many different languages might be interleaved.

Specifically, suppose that the linguistic input, coded as a binary sequence, $x$, is generated by a computable probability distribution, $\mu_C(x)$.[7] Intuitively, we can view this as



meaning that there is a computer program, *C*, (which might, for example, encode a grammar, as above) which receives random input *y*, from a stream of coin flips. When fed to C, this random input generates *x* as output, i.e., *C*(*y*) = *x*. The probability of this *y* is $2^{-l(y)}$ (the probability of generating any specific binary sequence of length *l*(*y*) from unbiased coin flips). Many *y* may generate the same *x*, so the probability of an output with initial segment *x*, $\mu_C(x)$, is the sum of the probabilities of such *y*:

$$\mu_C(x) = \sum_{y:C(y)=x_-} 2^{-l(y)} \qquad (1)$$

The distribution $\mu_C(x)$ is built on a simplicity principle: outputs which correspond to short programs for the computer program, *C*, are overwhelmingly more probable than outputs for which there are no short programs.

The learner's task, then, can be viewed as approximating $\mu_C(x)$, given a sample *x*, generated from the computer program, *C*. So, for example, if *C* generated independent samples from a specific stochastic phrase structure grammar, then the learner's aim is to find a probability distribution which matches the probabilities generated by that stochastic grammar as accurately as possible. To the extent that this is possible, we might conjecture that the learner should (i) be able to predict how the corpus will continue; (ii) decide which strings are allowed by $\mu_C(x)$; and (iii) generate output similar to that generated by $\mu_C(x)$. Framing these points in terms of language acquisition, this means that, by approximating $\mu_C(x)$, the learner can, to some approximation, (i) predict what phoneme, word, or sentence will come next (insofar as this is predictable at all); (ii) learn to judge grammaticality; and (iii) learn to produce language, indistinguishable from that to which it has been exposed. We explore these issues in turn in Sections 3-5.



How, then, can the learner approximate $\mu_C(x)$, given that it has exposure to just one (admittedly arbitrarily long) corpus $x$, and no prior knowledge of the specific computational process, $C$, which has generated this corpus? It turns out that we can make progress by assuming only that the learner can, in principle, entertain all and only computable hypotheses—i.e., that the learner's representational resources are *universal*: i.e., sufficient to encode any possible computation. Elementary results in computability theory (e.g., Odifreddi, 1988) have shown that this assumption of universality is surprisingly mild, and is satisfied by very simple abstract languages (such as the lambda calculus, Barendregt, 1984) and familiar practical languages from Fortran, to C++, to Java and Scheme. We assume, then, that the brain (and our ideal learner) has at least these representational resources.

We have stated that a simplicity-based learner favor simple "explanations," measured in terms of code length in some programming language. But surely the length of a program depends on the programming language used? What may be easy to write in Matlab may be difficult to write in Prolog; and vice versa. It turns out, though, that the choice of programming language affects program lengths only to a limited degree. An important result, known as the invariance theorem (Li & Vitányi, 1997), states that, for any two universal programming languages, the length of the shortest program for any computable object in each language is bounded by a fixed constant. A caveat is appropriate, however: "invariance" up to an additive constant is sufficient for establishing mathematical results, such as those below; but choice of representation language is crucial for making learning practically feasible, as we shall note in Section 7.[8] Nonetheless, so long as we assume that the learner's coding language is universal, we can avoid having to provide a specific account of the program that the learner uses.[9]



Now suppose the learner assumes only that the corpus, $x$, is generated by a computable process (and hence makes no assumptions that it is generated by a specific type of grammar, or indeed, any grammar at all; makes no assumption that "sentences" are sampled independently, etc.). Then the probability of each possible $x$ is given by the probability that this sequence will be generating from the output of a random input, $y$, of length $l(y)$ (as before, by random coin flips) fed to a *universal* computer, $U$.[10] Analogous to (1), we can define this "universal monotone distribution" (Solomonoff, 1978) $\lambda(x)$:

$$\lambda(x) = \sum_{y:U(y\ldots)=x} 2^{-l(y)} \qquad (2)$$

where $U(y)$ are programs $y$ written in the universal programming language. Thus, an ideal learner draws on its universal programming language and the simplicity principle to formulate $\lambda(x)$. Remarkably, it turns out that $\lambda(x)$ serves as a good enough approximation to $\mu_C(x)$ to allow the ideal learner to predict future linguistic input; and we show below that this allows the ideal learner to make grammaticality judgments, produce grammatical utterances, and map sound to meaning.

What is the mysterious $\lambda(x)$ in more concrete terms? Roughly, it is what would result from randomly typing into a computer; feeding the resulting "programs" (most of which will, of course, not even be syntactically valid, or will loop indefinitely) to the interpreter for some universal programming language (say, C++); and considering the outputs of the (small number of) valid and terminating programs. Thus, we can alter the familiar image of monkeys randomly hitting the keys on a typewriter and, supposedly, eventually generating the works of Shakespeare, to the image of monkeys typing *computer programs*, and generating outputs $x$ according to $\lambda(x)$. The probability $\lambda(x)$ will depend, of course, on the length of the shortest



program generating $x$, as short programs are overwhelmingly more likely to be chanced upon by the monkey.

We shall explore the remarkable properties of $\lambda(x)$ shortly. But it is worth noting at the outset that $\lambda(x)$ is known to be uncomputable (Li & Vitányi, 1997), and hence must be approximated. It remains an open question how closely $\lambda(x)$ can be approximated and how this affects learnability results.  Promisingly, computable approximations to the universal distribution can be developed into practical tools in statistics and machine learning (e.g., Rissanen, 1987; Wallace & Freeman, 1987). Related approximations will be considered briefly in Section 7 in relation to developing a methodology for assessing the learnability of specific linguistic patterns.

*3. Prediction*

One indication of the degree to which a learner understands the patterns in the data in any domain, is its ability to predict. Thus, if the linguistic input is governed by grammatical or other principles of whatever complexity, any learner that can predict how the linguistic material will continue, arbitrarily well, must, in some sense, have learned such regularities. Prediction has been used as a measure of how far the structure of a language has been learned since Shannon (1951); and is widely used as a measure of learning in connectionist models of language processing (Christiansen & Chater, 1994, 1999; Elman, 1990) . And, as we have noted, this result for prediction will be a foundation for results concerning grammaticality judgments, language production, and form-meaning mapping, as we discuss in subsequent sections.



We formulate the task of prediction as follows. At each point in a binary sequence $x$ (encoding our linguistic input), generated by computer $C$, the probabilities that, given input $x$, that the next symbol is 0 or 1 can be written:

$$\mu_C(0 \mid x) = \frac{\mu_C(x0)}{\mu_C(x)} \; ; \qquad \mu_C(1 \mid x) = \frac{\mu_C(x1)}{\mu_C(x)} \qquad\qquad (3)$$

where $\mu_C(0 \mid x)$ and $\mu_C(1 \mid x)$ represent the probabilities that the subsequence $x$ is followed by a 0 and 1 respectively; and $\mu_C(x0)$ and $\mu_C(x1)$ are the probabilities the specific sequence of $x$ followed by 0 or 1, respectively.  But the ideal learner does not have access to $\mu_C(x)$, but instead uses $\lambda(x)$ for prediction. Thus, the learner's predictions for the next item of a binary sequence that has started with $x$ is:

$$\lambda(0 \mid x) = \frac{\lambda(x0)}{\lambda(x)}; \qquad \lambda(1 \mid x) = \frac{\lambda(x1)}{\lambda(x)} \qquad\qquad (4)$$

A key result by Solomonoff (1978), which we call the *Prediction Theorem*, shows that, in a specific rigorous sense, the universal monotone distribution $\lambda$, described above, is reliable for predicting *any* computable monotone distribution, $\mu$, with very little expected error. More specifically, the difference in these predictions is measured by the square of difference in the probabilities that $\mu$ and $\lambda$ assign to 0 being the next symbol:

$$\text{Error}(x) = \left( \lambda(0 \mid x) - \mu(0 \mid x) \right)^2 \qquad\qquad (5)$$



And the *expected* sum-squared error for the *n*th item in the sequence is:

$$s_n = \sum_{x:l(x)=n-1} \mu(x)\text{Error}(x) \qquad (6)$$

The better $\lambda$ predicts $\mu$, the smaller $s_n$ will be. Given this, the overall expected predictive success of the method across the entire sequence is obtained by summing the $s_n$ across all *n*:

$$\sum_{n=1}^{\infty} s_n \qquad (7)$$

Solomonoff's Prediction Theorem shows that predictions using $\lambda$ approximate any computable distribution, $\mu$, so that $\sum_{n=1}^{\infty} s_n$ is bounded by a constant. Thus, as the amount of data increases, the expected prediction error goes to 0.  Specifically, the following result holds:

**Prediction Theorem (Solomonoff, 1978):** Let $\mu$ be a computable monotone distribution, predicted by a universal distribution $\lambda$. Then,

$$\sum_{n=1}^{\infty} s_n \leq \frac{\log_e 2}{2} K(\mu) \qquad (8)$$

where $K(\mu)$ is the length of the shortest program on the universal machine that implements $\mu$, known as its Kolmogorov complexity  (see Li & Vitányi, 1997, for further details, and an accessible proof).

The Prediction Theorem shows that learning by simplicity can, in principle, be expected to converge to the correct conditional probabilities for predicting subsequent linguistic material. This implies that the learner is able to learn the structure of the language---because if



not, the learner will not know which sentences are likely to be said, and hence will make prediction errors. This results suggest that, given sufficient positive evidence, linguistic restrictions, such as those on the allowed contraction of *is* mentioned above, are learnable from positive evidence. Here "sufficient" means enough language input has been observed such that the (more complex) grammar which contains the restriction provides the simplest overall coding of the data, because it provides an efficient specification of that input. The learner which does not learn these restrictions will continue to predict the ungrammatical form when it is not allowed, and thus accrue an infinite number of prediction errors. Note that while the Prediction Theorem demonstrates that an ideal learner, with sufficient positive evidence, will learn to respect these linguistic restrictions, there is no claim that the learner can recover grammar that generated the language---but the learner's predictions will capture the structure of the language arbitrarily closely.

## *4. Learning grammatical judgments*

One of the distinctive shifts from Bloomfield's (1933) version of structural linguistics to Chomsky's (1957) generative grammar concerns methodology: while Bloomfield considered the goal of linguistics to be inducing patterns in language from corpora of utterances, Chomsky rejected this approach, and stressed instead capturing native speaker intuitions about, for example, the grammaticality of sentences. Our discussion of prediction, based on the linguistic input to the learner, seems closely allied to Bloomfield's perspective. But Chomsky's approach presents a fresh challenge: human language learners appear not just to learn to predict, based on the structure of what they hear---instead, people appear to be able learn to be able to distinguish grammatical from ungrammatical sentences from positive evidence alone. This



raises the question: under what conditions are grammaticality judgements learnable from positive data?

It turns out that the task of prediction naturally extends surprisingly naturally to that of grammaticality judgments. The crucial move is to consider predictions for larger units linguistic material (e.g., words, rather than binary codes) and ask how often the predicted utterance will correspond to a continuation that *is* a grammatical sentence. The crucial question is how far the learner's predictions fit with the set of options that are grammatically possible in the language. Specifically, we can ask: How often does the learner *overgeneralize* such that its guesses violate the rules of the language (e.g., predicting a contraction of *is* where it is not allowed)? And, conversely, how often does the learner *undergeneralize* what is possible, such that it fails to guess continuations that are acceptable (e.g., not predicting a contraction when it *is* allowed)? Results for overgeneralization and undergeneralization errors are examined in turn.

### 4.1 Grammaticality errors: overgeneralization

When considering grammaticality, it is, as we have noted, convenient to consider language input as a sequence of words, rather than coded as a binary form. Thus, instead of dealing with distributions, $\mu$, $\lambda$, over binary sequences, one may consider distributions $P_\mu$ and $P_\lambda$.over sequences of a finite vocabulary of words. Suppose that the learner has seen a corpus, $x$, of *j-1* words and has a probability $\Delta_j(x)$ of incorrectly guessing a *j*th word which happens to be ungrammatical, i.e., the string cannot be completed to produce a grammatical sentence. One can write:



$$\Delta_j(x) = \sum_{\substack{k:xk \text{ is un grammatical,} \\ l(x)=j-1}} P_\lambda(k \mid x) \tag{9}$$

As before, we focus on the *expected* value $\langle \Delta_j \rangle$:

$$\langle \Delta_j \rangle = \sum_{x:l(x)=j-1} P_\mu(x)\Delta_j(x) \tag{10}$$

This expected value captures the probability that the learner's prediction concerning the *j*th word will not actually be allowable in the language—that the learner overgeneralizes what the language contains. But such overgeneralizations are, of course, a failure of prediction---and we know, from the Prediction Theorem above, that errors in the learner's predictions are gradually eliminated. So the Prediction Theorem can be used to provide a 'bound' on the number of overgeneralization errors that the learner will generated. Specifically, it is possible to derive the following 'overgeneralization theorem' (Chater & Vitányi, 2007):

$$\sum_{j=1}^{\infty} \langle \Delta_j \rangle \le \frac{K(\mu)}{\log_e 2} \tag{11}$$

That is, the total expected amount of probability devoted by the learner to overgeneralizations, in the course of encountering an infinite corpus, sums to a finite quantity; and this quantity is close to the length of the shortest program that generates the linguistic data. Thus, the expected amount of overgeneralization must tend to zero, as more of the corpus has been encountered; and the number of errors will depend on the complexity of the language to be learned (where complexity is measured in terms of program length).

The ability to deal with overgeneralization of the grammar from linguistic experience is particularly relevant to previous discussions of the "logical problem" of language learnability, discussed above (Baker & McCarthy, 1981; Hornstein & Lightfoot, 1981; Pinker, 1979; Pinker, 1984). The learner only hears a finite corpus of sentences. Assuming the language is



infinite, a successful learner must therefore infer the acceptability of an infinite number of sentences that it has never heard. Thus, *not* hearing a sentence cannot be evidence against its existence. As noted above, this has raised the puzzle of whether it is possible for overly general grammars ever to be eliminated by the learner. The overgeneralization theorem shows that an ideal learner using the simplicity principle will eliminate overly general grammars, given a sufficiently large corpus.

*4.2 Grammaticality errors: undergeneralization*

The universal distribution used by the ideal learner was defined as being a combination of all possible (computable) distributions over corpora,  and thus all grammatical sentence in the language will always be deemed possible (assigned non-zero probability).  This immediately implies that an ideal learner will never strictly undergeneralize, i.e., incorrectly deem a grammatical utterance to have probability 0. But perhaps an ideal learner could drastically *underestimate* a sentence's probability of occurrence.  One can investigate the extent to which an ideal learner might commit such errors of 'soft' undergeneralization, putting an upper bound on the number of soft undergeneralizations an ideal learner will make. Suppose that the learner underestimates, by a factor of at least *f*, the probability that word *k* will occur after linguistic material *x*. That is, $P_\lambda(k|x) f \leq P_\mu(k|x)$. Let $\Lambda_{j,f}(x)$ denote the probability that the word that is the true continuation will be one of the *k* for which this underestimation occurs:

$$\Lambda_{j,f}(x) = \sum_{k:f.P_\lambda(k|x)\leq P_\mu(k|x)} P_\mu(k\mid x) \qquad (12)$$

The corresponding *expected* probability is:

$$\left\langle \Lambda_{j,f} \right\rangle = \sum_{x:l(x)=j-1} P_\mu(x)\Lambda_j(x) \qquad (13)$$



Then, the following undergeneralization theorem holds, which bounds the expected number of undergeneralization errors throughout the corpus, i.e., $\sum_{j=1}^{\infty}\left\langle \Lambda_{j,f} \right\rangle$:

$$\sum_{j=1}^{\infty}\left\langle \Lambda_{j,f} \right\rangle \leq K(\mu)\frac{1}{\log_2 f/e} \qquad (14)$$

so long as $f > e$, where $e$ is the mathematical constant 2.71...

Thus the expected number of 'soft' undergeneralizations is bounded, even for an infinitely long sequence of linguistic input and the expected rate at which such errors occur converges to zero. As with overgeneralizations, the upper bound is proportional to $K(\mu)$, the complexity of the underlying computational mechanism generating the language (including, presumably, the grammar). The higher the underestimation factor $f$ to be, the fewer such undergeneralizations occur.

In summary, formal results have shown that an ideal learner, using the universal probability distribution, $P_\lambda$, and derived from the simplicity principle, can learn to make accurate grammaticality judgments that avoid both overgeneralizations and undergeneralizations---an issue that, as noted above, has been viewed as of fundamental importance in recent linguistic theorizing. In the description above, grammaticality judgments have been framed as the process of guessing which *word* comes next. However, it is important to note that these results extend to all other units of linguistic analysis, e.g., prediction of utterances on the level phonemes, syllables, or sentences.

## 5. Learning to Produce Language

One method of describing language production is to assume that it is simply a matter of predicting future utterances of arbitrarily long lengths. Thus, a learner, given an entire history



of linguistic input, can eventually "join in" and starts *saying* its predictions.  Production

success can be assessed by how well these productions blend in with the linguistic input --i.e.,

how well the learner's productions match those that other speakers of the language (i.e., those

producing the learner's corpus) might equally well have said. This is, of course, a highly

limited linguistic goal, given that a key purpose of language is to express one's own thoughts,

which may be diverge from what others have said before. (We will consider how this limitation

can partially be dealt with in the next section.) However, as a first step, one can begin to assess

a learner's ability to speak a language by assessing whether the learner can blend into the on-

going "conversation."

Blending in can be described as the ability to match the actual probability that a new

sequence of utterances, $y$, will follow the previous utterances, $x$, which have been heard so far

in the conversation.   This is the probability $\mu(y|x)$, which reflects the distribution of continued

sequences that would be uttered by speakers of the language. As before, the learner's stream of

utterances can be defined on any linguistic level, e.g., phonemes, words or sentences. Because

the ideal learner generates utterances using the distribution it learned in prediction, $\lambda$, the

learner will predict continuations according to $\lambda(y|x)$. The learner will blend in, to the extent

that $\lambda(y|x)$ is a good approximation to $\mu(y|x)$--i.e., the extent to which the learner has a

propensity to produce language that other speakers have a propensity to produce.  Note,

though, that the objective is now not merely predicting the next binary code, piecemeal; the

material to be predicted, $y$, can be an arbitrarily large chunk of linguistic material (e.g., an

entire clause or sentence).

It turns out  that $\lambda(y|x)$ is a good approximation to any relevant $\mu(y|x)$ (Li & Vitányi,

1997;  this result does not follow directly from the Prediction Theorem): If $\mu$ is, as above, a



probability distribution associated with a monotone computable process, and $\lambda$ denotes the universal distribution, then for any finite sequence $y$, as the length of sequence $x$ tends to infinity:

$$\frac{\lambda(y \mid x)}{\mu(y \mid x)} \to 1 \qquad (15)$$

with a probability tending to 1, for fixed utterance $y$ and growing prior linguistic experience $x$. Thus, viewing (15) in the context of language production, this means that, in the asymptote, the learner will blend in arbitrarily well, so that its language productions are indistinguishable from those of the language community to which it has been exposed.

## 6. Learning to map linguistic forms to semantic representations

In addition to being able to predict, make grammatical judgments, and produce linguistic regularities, language acquisition also involves associating linguistic forms with *meanings*. Indeed, to the ability to judge grammaticality, or produce language indistinguishable from that of one's speech community, would be pointless unless it were associated with the ability to communicate: to map from utterances to some representation of their interpretations, and back (we remain neutral here about nature of these representations).

A common assumption among researchers (Pinker, 1989) is that the child can infer semantic interpretations from linguistic context. Therefore the problem of learning interpretations from linguistic input can be framed as a problem of induction from pairs of linguistic and semantic representations. One can then show that, given sufficient pairs, the ideal learner is able to learn this mapping, in either direction, in a probabilistic sense. This result holds even though the mapping between linguistic and semantic representations can be many-



to-many. That is, linguistic representations are often ambiguous; and the same meaning can often be expressed linguistically in a number of different ways.

Concretely, we view the learner's problem as learning a relation between linguistic representations (e.g., as the $i^{th}$ string of words), $S_i$, and a semantic interpretations, $I_j$, (representing the $j^{th}$ meaning of the string). Suppose that the language consists of a set of ordered pairs $\{<S_i, I_j>\}$, which we sample randomly and independently according to computable probability distribution $\Pr(S_i, I_j)$.

Now we can apply the Prediction Theorem, as described above, but where the data now consist of pairs of sentences and interpretation, rather than strings of phonemes or words. So, when provided with a stream of sentence-interpretation pairs sampled from $\Pr(S_i, I_j)$, the learner can, to some approximation, infer the joint distribution $\Pr(S_i, I_j)$. But, of course, approximating this joint distribution is only possible if the learner can approximate the relationship between sentences $S_i$ and interpretations $I_j$.

Writing the length of the shortest program that will generate the computable joint distribution, $\Pr(S_i, I_j)$, as  $K(\Pr(S_i, I_j))$, the Prediction Theorem above ensures that this joint distribution is learnable from positive data by an ideal learner---if that positive data includes both form and meaning. Specifically, by (8), this has an expected sum-squared error bound of $\frac{\log_2 2}{2} K(\Pr(S_i, I_j))$. Hence the expected value of error per data sample, will tend to zero because this bound is finite, but the data continues indefinitely.

If ordered pairs of $<S_i, I_j>$ items can be predicted, then the relation between sentences and interpretations can be captured; and this implies that the mapping from sentences to probabilities of interpretations of those sentences, $Pr(I_j | S_i)$, and the mapping from interpretations to probabilities of sentences with those interpretations, $Pr(S_i | I_j)$, are learnable.[11]



Thus, we can conclude that the ideal learner is able to learn to map back and forth between sentences and their interpretations, given a sufficiently large supply of sentence-interpretation pairs as data. That is, in this specific setting at least, the relation between form and meaning can be derived from positive data alone.

### 7. Scaling down simplicity: A practical method for assessing learnability

We have described a range of rather abstract theoretical results concerning the viability of language learning by simplicity. But how far can the simplicity-based approach be "scaled-down" to inspire concrete models of learning? The practical instantiation of the simplicity principle has been embodied using the minimum description length (MDL, Rissanen, 1987) and minimum message length (MML, Wallace & Freeman, 1987) frameworks. Simplicity has also widely been explored as general principle underpinning concrete models in a range of areas of perception and cognition (e.g., Attneave & Frost, 1969; Jacob Feldman, 2000; Hochberg & McAlister, 1953; Leeuwenberg, 1969), including language (e.g., Brent & Cartwright, 1996; Dowman, 2000; Ellison, 1992; Goldsmith, 2001; Onnis, Roberts & Chater, 2002; Vousden, Ellefson, Solity & Chater, 2011; Wolff, 1988). Closely related Bayesian methods have also been widely employed (e.g., Kemp, Perfors & Tenenbaum, 2007; Langley & Stromsten, 2000; Perfors, Regier & Tenenbaum, 2006; Stolcke, 1994).

The type of theoretical analysis that we have outlined above applies, by the invariance theorem, irrespective of specific choices of representations (as long as these are sufficiently powerful). But to make the approach computationally concrete requires choosing a specific representation—typically this will be a representational formalism developed in linguistics (e.g., some type of grammar). A code length can then be assigned both to the rules of the



grammar as well as to the corpus when encoded in terms of those rules (the corpus might consist of all the utterances that a learner has experienced so far, or a subset of these).

Suppose that we wish to evaluate how much data is required to learn a particular linguistic regularity. This can be heuristically assessed by comparing two grammars which are identical aside from the fact that only one captures the regularity of interest. For example, consider how we might assess whether the corpus contains sufficient information to learn the restrictions on cases where *is* can be contracted that we described earlier. A grammar containing this additional regularity requires, of course, greater code-length than one that does not; but, on the other hand, because the resulting model of the language is more accurate, the code length of the corpus, given this more accurate model, will be shorter. Whether the 'balance' favors the more complex but accurate grammar (thus allowing the restrictions on contraction to be learned) depends on the corpus. For a null, or a short, corpus, the advantage of a more accurate language model will not be sufficient; however, once the corpus becomes sufficiently long, the more accurate model will produce a shorter overall code-length, and the regularity will be learned. The question is: how long does the corpus need to be, for the regularity to learnable?

As discussed in Section 1, the simplicity principle automatically trades-off competing simpler and complex grammars. Simple, but over-general, grammars can be described more briefly, but because they are less accurate descriptions of actual language structure, they give an inefficient descriptions of language input. More complex grammars, which include linguistic restrictions, have more complex descriptions, but better capture the language and so give more efficient descriptions of the language input. By "investing" in a more complicated grammar, which contains a restriction on a construction, the language speaker obtains encoding



savings every time the construction occurs. Intuitively, a linguistic restriction is learned when the relevant linguistic context occurs often enough that the accumulated savings makes the more complicated grammar worthwhile, just as the extra cost of an energy saving appliance is justified if it is used sufficiently often.

Recently, a simple and practical framework for assessing learnability of a wide variety of linguistic constructions under simplicity has been proposed (Hsu and Chater, 2010). Using natural-language corpora to simulate the language input available to the learner, this framework quantifies learnability (e.g., in estimated number of years of linguistic exposure) for any given linguistic constraint, such as the contraction of *is* mentioned earlier.

To get started, we need some description of the grammatical rule to be learned, i.e., a description of an original, incorrect (over-general) grammar and the new, correct grammar, which contains the restriction rule. Moreover, we need a corpus which will serve as a proxy for the learner's input. Given these, the framework provides a method for quantifying an upper bound on learnability from language input. This framework assumes an ideal statistical learner and thus provides an upper bound on learnability based on language statistics. However, measures of learnability should give an indication of the ease with which various linguistic constraints can be learned.

While the details of implementing this framework are described elsewhere (Hsu & Chater, 2010; Hsu, Chater & Vitányi, 2011), an intuitive description of how this framework works is as follows: Under this framework, the learnability is affected by three factors. (1) The first is the complexity of the rule to be learned (greater complexity will decreases learnability). (2) The second concerns with what probability the "disallowed forms" would otherwise be expected to appear in place of other similar constructions which do occur (e.g., how often do



non-contracted forms "he is," "she is" etc., and in which syntactic contexts).  (3) How frequently does the putative regularity apply in real language input. (1) and (2) determine how many occurrences of contexts where the regularity applies are needed for learning and (3) then will determine how many millions of words (or years of language input) are required to accrue the number of occurrences needed. These assumptions are all, of course, provisional; and hence results from this approach are suggestive rather than definitive.

Hsu and Chater (2010) applied this general framework to consider the learnability of various linguistic restrictions, many of which have been viewed as presenting fundamental learnability challenges. They assuming that a learner's input can be approximated using corpora of adult speech and writing, such as the Corpus of Contemporary American English (COCA). They found that the number of years of linguistic input required to learn putatively "unlearnable" constructions varied surprisingly widely, from a matter of months to more than a lifetime.

Might these learnability differences across different linguistic restrictions correlate with how well people actually do learn them? This was tested in an experiment on adult native English speakers in Hsu, Chater and Vitányi (2011). Figure 1a shows the predictions for 15 constructions from Hsu and Chater (2010), sorted in descending learnability. Figure 1b shows how often the constructions occur per year of linguistic input, estimated from COCA. Note that the occurrence rates do not monotonically decrease with the years required to learn the construction, because other factors that affect learnability, e.g., (1) and 2) listed above). Interestingly, the more learnable the constraints according to the simplicity analysis, the better they are learned in practice by native speakers: as log(1/predicted years needed) increased, the difference in the grammatical acceptability of the grammatical vs. ungrammatical form of the



construction also increased. Thus, a simplicity-based approach to language acquisition can provide not only general learnability results, but concrete predictions concerning how people learn language.

## 8. Conclusion

In this paper, we have reviewed some recent results concerning learning language from experience by employing the simplicity principle: that is, favouring models of the language to the degree that they provide a short encoding of the linguistic input. We have shown theoretical results that indicate that an "ideal learner" implementing the simplicity principle can learn to predict language from experience; to determine which sentences of a language are grammatical to an arbitrarily good approximation (assuming, somewhat unrealistically, that the corpus of linguistic experience is noise-free, i.e., containing only grammatical sentences); to produce language; and to map between sentences and their interpretations. This "ideal" learning approach is valuable for determining what information is contained in a corpus. Yet it cannot be implemented computationally, as the relevant calculations are known to be uncomputable (Li & Vitányi, 1997). Nonetheless, we have also shown how a local approximation to such calculations can be used to choose between different grammars which do or do not contain specific regularities (especially those concerned with exceptions) that have been viewed as posing particular problems for theories of language acquisition. Overall, these results form part of a wider tradition of analytic and computational results on language learning which suggest that general *a priori* arguments about whether language acquisition requires language-specific innate constraints can be replaced by a more precise formal and empirical analysis.





Figure 1:

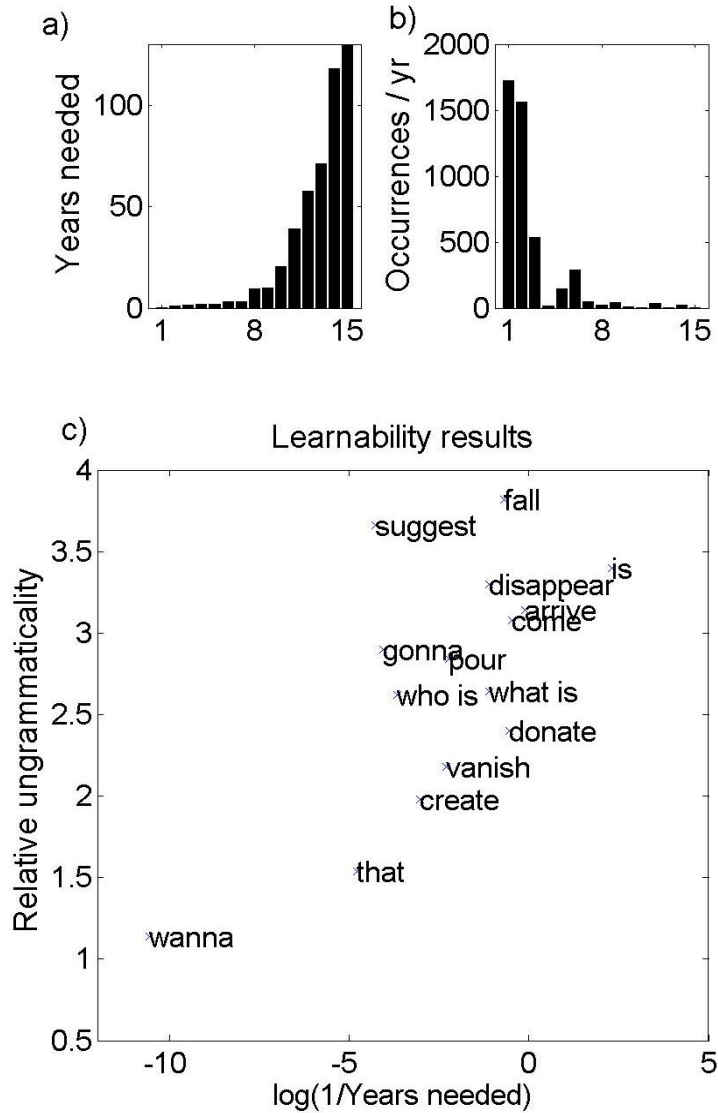

Figure 1: Learnability predictions and experimental evidence: These results are re-plotted from Hsu, Chater & Vitányi (2011). (a) Estimated years required to learn construction. (b) Number of occurrences per year (estimated from COCA). (c) Relative grammaticality vs. learnability for Sentence Set 1 (r = 0.67;p = 0.006). Relative grammaticality judgments were elicited from 200 native English speakers in an online study. Learnability is log of the inverse of the number of estimated years needed to learn the construction.



Reference List


Angluin, D. (1980). Inductive inference of formal languages from positive data. *Information and Control*, *45*, 117–135.

Angluin, D. (1988). Identifying languages from stochastic examples. Technical Report. Department of Computer Science, Yale University.

Attneave, E, & Frost, R. (1969). The determination of perceived tridimensional orientation by minimum criteria. *Perception & Psychophysics, 6,* 391-396.

Baker, C. L. & McCarthy, J. J. (1981). *The logical problem of language acquisition.* Cambridge, Mass: MIT Press.

Barendregt, H. P. (1984). *The lambda calculus*. Amsterdam: Elsevier.

Bloomfield, L. (1933). *Language*. New York: Henry Holt.

Bowerman, M. (1988). The 'No Negative Evidence' Problem: How do Children avoid constructing an overly general grammar? In J.Hawkins (Ed.), *Explaining Language Universals* (pp. 73-101). Oxford: Blackwell.

Brent, M. R. & Cartwright, T. A. (1996). Distributional regularity and phonotactic constraints are useful for segmentation. *Cognition*, *61*, 93-126.

Brown, R. & Hanlon, C. (1970). *Derivational complexity and order of acquisition in child speech*. (J.R. Hayes ed.) New York: Wiley.





Carlucci & Case, J. (2013). On the necessity of U-shaped learning. *Topics in Cognitive Science* (this issue).

Chater, N. & Vitányi, P. (2002). Simplicity: A unifying principle in cognitive science? *Trends in Cognitive Sciences*, *7*, 19–22.

Chater, N. & Vitányi, P. (2007). Ideal learning' of natural language: positive results about learning from positive evidence. *Journal of Mathematical Psychology, 51,* 135-163.

Chomsky, N. (1957). *Syntactic structures*, The Hague/Paris: Mouton.

Chomsky, N. (1980). *Rules and representations*. Cambridge, MA: MIT Press.

Christiansen, M. H. & Chater, N. (1994). Generalization and connectionist language learning. *Mind & Language, 9,* 273-287.

Christiansen, M. H. & Chater, N. (1999). Connectionist natural language processing: The state of the art. *Cognitive Science, 23,* 417-437.

Christiansen, M. H. & Chater, N. (2007). Generalization and connectionist language learning. *Mind & Language, 9,* 273-287.

Christiansen, M. & Chater, N. (2010). Language acquisition meets language evolution. *Cognitive Science*, *34*, 1131-1157.

Clark, A. & Eyraud. R. (2007). Polynomial identification in the limit of substitutable context-free languages. *Journal of Machine Learning Research*, 8, 1725-1745.





Clark, A., & Lappin, S. (2013). Complexity in language acquisition. *Topics in Cognitive Science* (this issue).

Cover, T. M. & Thomas, J. A. (2006). *Elements of Information Theory (2nd Edition)*, Wiley.

Crain, S. & Lillo-Martin, D. (1999). *Linguistic theory and language acquisition*. Oxford: Blackwell.

Demetras, M., Post, K., & Snow, C. (1986). Feedback to first language learners: The role of repetitions and clarification questions. *Journal of Child Language*, *13*, 275-292.

Dowman, M. (2000). Addressing the learnability of verb subcategorizations with Bayesian inference. In L. R. Gleitman & A. K. Joshi (Eds.). *Proceedings of the Twenty Second Annual Conference of the Cognitive Science Society*. Mahwah, NJ: Erlbaum.

Dresher, B. & Hornstein, N. (1976). On Some Supposed Contributions of Artificial Intelligence to the Scientific Study of Language. *Cognition, 4,* 321-398.

Ellison, M. (1992). The machine learning of phonological structure. PhD Thesis, University of Western Australia.

Elman, J. (1990). Finding Structure in Time. *Cognitive Science, 14,* 179-211.

Feldman, Jacob (2000). Minimization of boolean complexity in human concept learning. *Nature, 403,* 630-633.

Feldman, Jerome (1972). Some decidability results on grammatical inference and complexity. *Information and Control*, *20*, 244–262.





Gold, E. M. (1967). Language identification in the limit. *Information and Control*, *16*, 447–474.

Goldsmith, J. (2001). Unsupervised learning of the morphology of a natural language. *Computational Linguistics, 27,* 153-198.

Hochberg, J. & McAlister, E. (1953). A quantitative approach to figure "goodness." *Journal of Experimental Psychology*, *46*, 361-364.

Horning, J. J. (1969). A study of grammatical inference. Technical Report CS 139, Computer Science Department, Stanford University.

Hornstein, N. & Lightfoot, D. (1981). *Explanation in linguistics: the logical problem of language acquisition.* London: Longman.

Hsu, A. & Chater, N. (2010). The logical problem of language acquisition: A probabilistic perspective. *Cognitive Science, 34,* 972-1016.

Hsu, A., Chater, N., & Vitányi, P. (2011). The probabilistic analysis of language acquisition: Theoretical, computational, and experimental analysis. *Cognition, 120,* 380-390.

Jain, S., Osherson, D. N., Royer, J. S., & Kumar Sharma, A. (1999). *Systems that learn (2nd edition).* Cambridge, MA: MIT Press.

Kemp, C., Perfors, A., & Tenenbaum, J. B. (2007). Learning overhypothesis with hierarchical Bayesian models. *Developmental Science, 10,* 307-321.

Klein, D. & Manning, C. (2005). Natural language grammar induction with a generative constituent-context model. *Pattern Recognition, 38,* 1407-1409.




Langley, P. & Stromsten, S. (2000). Learning context-free grammars with a simplicity bias. *Proceedings of the Eleventh European Conference on Machine Learning,* 220-228.

Leeuwenberg, E. L. J. (1969). Quantitative specification of information in sequential patterns. *Psychological Review, 76,* 216-220.

Li, M. & Vitányi, P. (1997). *An introduction to Kolmogorov complexity and its applications* (2$^{nd}$ Edition). Springer.

Mach, E. (1959). *The analysis of sensations and the relation of the physical to the psychical*. New York: Dover Publications (Original work published 1886).

MacWhinney, B. (1993). The (il)logical problem of language acquisition. In *Proceedings of the 15th annual conference of the Cognitive Science Society* (pp. 61–70). Mahwah, NJ: Erlbaum.

MacWhinney, B. (2004). A multiple process solution to the logical problem of language acquisition. *Journal of Child Language*, *31*, 883–914.

Marcus, G. F. (1993). Negative evidence in language acquisition. *Cognition, 46,* 53-85.

Niyogi, P. (2006). *The computational nature of language learning and evolution*. Cambridge, MA: MIT Press.

Odifreddi, P. (1988). *Classical recursion theory*. North Holland: Elsevier.

Onnis, L., Roberts, M., & Chater, N. (2002). Simplicity: A cure for overgeneralizations in language acquisition? *Proceedings of the 24th Annual Conference of the Cognitive Science Society,* 720-725.




Pereira, F. C. N. & Warren, D. H. D. (1980). Definite clause grammars for language analysis. *Artificial Intelligence*, *13*, 231-278.

Perfors, A., Regier, T., & Tenenbaum, J. B. (2006). Poverty of the Stimulus? A rational approach. *Proceedings of the Twenty-Eighth Annual Conference of the Cognitive Science Society,* 663-668.

Pinker, S. (1979). Formal models of language learning. *Cognition, 7,* 217-283.

Pinker, S. (1984). *Language learnability and language development*. (7 ed.) Cambridge, MA: Harvard University Press.

Pinker, S. (1989). *Learnability and Cognition: The acquisition of argument structure*. Cambridge, MA: MIT Press.

Pinker, S. & Bloom, P. (1990). Natural language and natural selection. *Behavioral and Brain Sciences, 13,* 707-784.

Pylyshyn, Z. W. (1984). *Computation and Cognition: Toward a Foundation for Cognitive Science*, Cambridge, MA: Bradford Books/MIT Press.

Rissanen, J. (1987). Stochastic complexity. *Journal of the Royal Statistical Society, Series B*, *49*, 223–239.

Rohde, D. L. T., & Plaut, D. C. (1999). Language acquisition in the absence of explicit negative evidence: How important is starting small? *Cognition*, *72*, 68–109.

Shannon, C. E. (1951). Prediction and entropy of printed English. *Bell Systems Technical Journal, 31,* 64.




Solomonoff, R. J. (1978). Complexity-based induction systems: comparisons and convergence theorems. *IEEE Transactions on Information Theory, IT, 24,* 422-432.

Steyvers, M., Griffiths, T., & Dennis, S. (2006). Probabilistic inference in human semantic memory. *Trends in Cognitive Sciences, 10,* 309-318.

Stolcke, A. (1994). *Bayesian Learning of Probabilistic Language Models.* Department of Electrical Engineering and Computer Science, University of California Berkeley.

Tomasello, M. (2004). Syntax or semantics? Response to Lidz et al. *Cognition*, *93*, 139–140.

Wallace, C. S., & Freeman, P. R. (1987). Estimation and inference by compact coding. *Journal of the Royal Statistical Society, Series B*, *49*, 240–251.

Wharton, R. M., (1974). Approximate language identification, *Information and Control*, 26, 236-255

Vousden, J.I., Ellefson, M.R., Solity, J.E., & Chater, N. (2011). Simplifying reading: Applying the simplicity principle to reading. *Cognitive Science*, *35*, 34-78.

Wolff, J. G. (1988). Learning syntax and meanings through optimisation and distributional analysis. In Y. Levy, I. M. Schlesinger & M. D. S. Braine (Eds.), *Categories and processes in language acquisition*, (pp. 179-215). Hillsdale, NJ: LEA.



Footnotes

1. Language acquisition involves dealing with other challenges, including the computational complexity of searching the space of grammars (Clark & Lappin, 2013), but these are outside our scope here.

2. Typical speech is, of course, full of grammatical errors, repetitions, and incomplete utterances. Along with most other learnability analyses, we will ignore the "noisy" character of linguistic input.

3. All code lengths are assumed, by convention, to be written in a binary alphabet.

4. This approach implicitly assumes, among other things, no sequential dependencies between sentences, but generalizations are relatively straightforward.

5. Recovery from overgeneralization can be explored in a number of frameworks, for example, Carlucci and Case (2013).

6. Informally, we can view this process as embodying a Turing machine (or any other computer) combined with a source of randomness (i.e., a sequence of coin flips). The source of randomness captures the possibility that the process of generating the linguistic input may be non-deterministic (although it need not be); the restriction to computable probability distributions requires that the *structure* in the linguistic input is computable.

7. Strictly, $\mu$ is a measure, rather than a probability distribution, as the sequence is infinite; indeed, it is actually a semi-measure. Measures and semi-measures are generalizations of the standard notion of probability distributions. We ignore these technicalities here (see Chater & Vitányi, 2007; and Li & Vitányi, 1997).

8. Note, though, that people presumably will *share* a mental representation language. Hence, the representational language used to formulate hypotheses in learning language will



presumably automatically be ideally suited to the natural languages that have been learned and generated by past generations of speakers (see, e.g., Christiansen & Chater, 2010).

9. We have discussed how (probabilistic) generative grammars and computer programs generate linguistic data. The relation between these can be very close: in some formalisms (e.g., Definite Clause Grammars, Pereira & Warren, 1980), the program generating from the grammar is just a specification of the grammar itself.  In general, the picture is slightly more complex, but we do not consider this further here.

10. Technically it is important that this computer is monotone (Chater & Vitányi, 2007), but we shall ignore this complication.

11. Of course, if interpretation, $I_j$, is such that $\Pr(I_j)=0$, then the fact that $\Pr(I_j, S_i)$ can be approximated arbitrarily well says nothing about $\Pr(S_i|I_j)$; similarly for sentences, $S_i$, such that $\Pr(S_i)=0$. But the learner presumably needs only learn sentences that express meanings that might actually arise; and interpret sentences that might actually be said, so this restriction is fairly mild.